\documentclass[10pt,twocolumn,letterpaper]{article}

\usepackage{iccv}              %

\usepackage[dvipsnames]{xcolor}
\usepackage{multirow}
\usepackage{colortbl}
\usepackage{algorithm2e}
\usepackage{xspace}
\usepackage{graphicx}
\usepackage{overpic}
\usepackage{wrapfig}
\usepackage{bm}
\usepackage{bbm}
\usepackage[normalem]{ulem}

\definecolor{lightgray}{gray}{0.9}
\definecolor{mediumgray}{gray}{0.7}

\usepackage{graphicx}
\usepackage{amsmath}
\usepackage{booktabs}

\usepackage{amssymb}%
\usepackage{pifont}%

\usepackage[accsupp]{axessibility}

\newcommand{\revi}[1]{{\color{black}#1}}
\newcommand{\reviICCV}[1]{{\color{black}#1}}
\newcommand{\parag}[1]{{\noindent\textbf{#1}}}

\newcommand{\name}{\emph{VQ-SGen}\xspace}

\definecolor{iccvblue}{rgb}{0.21,0.49,0.74}
\usepackage[pagebackref,breaklinks,colorlinks,citecolor=iccvblue]{hyperref}

\def\paperID{8396} %
\def\confName{ICCV}
\def\confYear{2025}

\title{\name: A Vector Quantized Stroke Representation for \\ Creative Sketch Generation}

\author{Jiawei Wang$^{1,2}$ \quad Zhiming Cui$^2$ \quad Changjian Li$^1$\\ [0.3em]
$^1$The University of Edinburgh \quad $^2$ShanghaiTech University 
\and
\href{https://enigma-li.github.io/projects/VQ-SGen/VQ-SGen.html}{https://enigma-li.github.io/projects/VQ-SGen/VQ-SGen.html}
}

\begin{document}
\maketitle

\begin{abstract}
This paper presents \name, a novel algorithm for high-quality \revi{creative} sketch generation.
Recent approaches have framed the task as pixel-based generation either as a whole or part-by-part, neglecting the intrinsic and contextual relationships among individual strokes, such as the shape and spatial positioning of both proximal and distant strokes.
To overcome these limitations, we propose treating each stroke within a sketch as an entity and introducing a vector-quantized (VQ) stroke representation for fine-grained sketch generation.
Our method follows a two-stage framework - in stage one, we decouple each stroke's shape and location information to ensure the VQ representation prioritizes stroke shape learning. In stage two, we feed the precise and compact representation into an auto-decoding Transformer to incorporate stroke semantics, positions, and shapes into the generation process. 
By utilizing tokenized stroke representation, our approach generates strokes with high fidelity and facilitates novel applications, such as \revi{text or class label conditioned generation and sketch completion}. %
Comprehensive experiments demonstrate our method surpasses existing state-of-the-art techniques on the CreativeSketch dataset, underscoring its effectiveness.
\end{abstract}
    
\section{Introduction}
\label{sec:intro}

\begin{figure}[!t]
    \centering
    \includegraphics[width=1.0\linewidth]{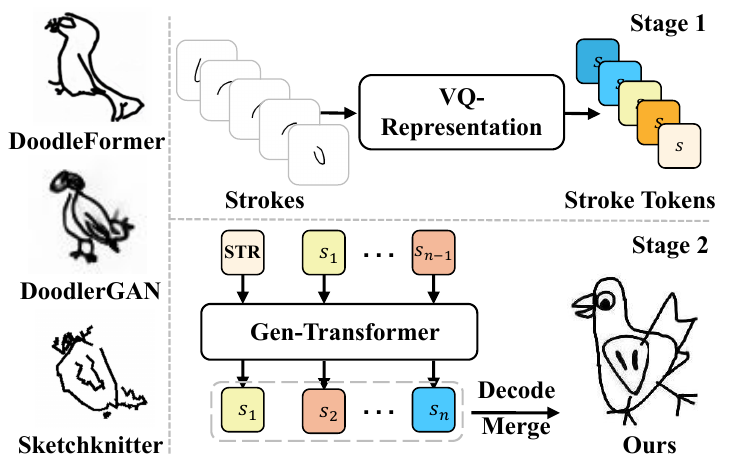}
    \vspace{-7mm}
    \caption{The goal of \revi{creative} sketch generation is to generate vivid sketches, 
    \eg, the birds from existing methods on the left. We have proposed \name, for high-quality sketch generation with a new vector-quantized (VQ) representation and an efficient generator. See one of our results on the right for comparison.
    }
    \vspace{-5mm}
    \label{pic:teaser}
\end{figure}

Sketches are an intuitive mechanism widely used by humans to communicate ideas, convey concepts, and express creativity. Over the past decades, significant research has been conducted on sketch-related tasks, including retrieval \cite{sain2021stylemeup, koley2024handle,sain2024freeview, Bandyopadhyay_2024_CVPR,Koley_2024_CVPR1}, semantic segmentation \cite{wang2024contextseg, zheng2024creativeseg}, and generation \cite{ge2020creative, bhunia2022doodleformer, wang2023sketchknitter, utintu2024sketchdeco, gao20243d,Koley_2024_CVPR}.
\revi{In this paper, we focus on \emph{creative sketch generation}, a novel task introduced by Ge \etal \cite{ge2020creative}. Unlike traditional sketch generation, this task aims to produce diverse, complex, and aesthetically appealing sketches, forming more imaginative depictions of familiar visual concepts, rather than generating conventional and typical representations of these concepts. 
This, thereby, places high demands on the generation capabilities of automated methods.}
\revi{
Benefiting from the advancement of diffusion models, the notable and relevant progress for sketch generation is Diffsketcher \cite{xing2023diffsketcher}, a novel approach for text-to-sketch generation. 
However, the large-scale backbone diffusion model is pre-trained on realistic images, and their resultant sketches thus have high fidelity and realism conforming to human imagination but fail to promote creativity.
Therefore, dedicated methods for creative sketch generation are needed.
}

Existing methods \cite{ge2020creative,bhunia2022doodleformer} tackle creative sketch generation by producing pixels either as a whole sketch or as segmented parts, overlooking inherent and contextual relationships among individual strokes, such as their shapes and relative positions of both nearby and distant strokes. 
As a result, their generated sketches often have blurry local regions, and scattered or isolated strokes (see the head of the bird from DoodleFormer and DoodlerGAN in \cref{pic:teaser}).
A more recent work, SketchKnitter \cite{wang2023sketchknitter}, formulated the problem of sketch generation as a stroke points rectification process, where diffusion models are leveraged to rearrange a set of stroke points from an initial chaotic state into coherent forms. However, without the concept of a stroke entity, this method performs poorly when generating complex creative sketches (see the bird in \cref{pic:teaser} from SketchKnitter). %

To address these limitations, we propose \reviICCV{a novel stroke-level sketch generation approach} that treats each stroke as an independent entity \revi{\cite{wang2024contextseg,qu2023sketchxai}} and encodes it using a novel vector-quantized (VQ) representation. This compact, discrete representation enhances the learning process of the generative model by efficiently capturing essential stroke shape variations while minimizing redundancy.
Despite the discrete nature of the representation, we observe semantic-aware clustering within the stroke code space, which lays an ideal foundation for the generator to sample new strokes. 
Furthermore, our bespoke generator incorporates the shape, semantic, and position into its generation process, effectively samples new sketches in the compressed VQ space, and accurately reasons out the structural and semantic relationship between strokes, resulting in sketches that are both appealing and structurally coherent.

Specifically, our method \name, adopts a two-stage framework. In the first stage, we decouple the shape and location information of each stroke. This decoupling allows the vector-quantized (VQ) representation to concentrate on capturing stroke shapes,  minimizing interference from positional data during the learning process. 
By prioritizing shape as the primary focus, the model learns a highly compact and informative, shape-oriented representation that preserves the inherent shape features of each stroke while reducing unnecessary redundancy. 
In the second stage, we employ an autoregressive Transformer to work with our VQ representation.
The Transformer progressively integrates the stroke's semantic (\ie, the category label), shape, and location into the generation process. 
Through the autoregressive decoding mechanism, the Transformer captures each stroke's semantic, shape, and context relationship with neighboring strokes, ensuring that the shape and positioning are accurate and harmoniously aligned with its semantic role within the overall composition. As a result, our approach effectively avoids all the artifacts observed in previous methods.

To demonstrate the effectiveness of our method, we have conducted experiments on the CreativeSketch~\cite{ge2020creative} dataset. Comprehensive experiments including comparisons and ablation studies show that our method surpasses existing state-of-the-art techniques. 
In summary, our main contributions are as follows:
\begin{itemize}
    \item  We propose \reviICCV{a novel stroke-level approach, \name, for the challenging creative sketch generation task.}  %
    \item We propose a new vector-quantized (VQ) stroke representation that treats each stroke as an entity, capturing essential shape features compactly and reducing redundancy to serve as the basis for stroke-level sketch generation.
    \item We propose a bespoke generator, leveraging the shape, semantic, and spatial location together, achieving high-quality generation. 
    \item We extensively evaluated our method, demonstrating superior performance. A user study shows that \name consistently performs favorably against SoTA methods.
\end{itemize}

\section{Related Work}
\parag{Sketch Representation Learning.}
Based on the data format of sketches, existing sketch representation learning methods can be broadly divided into three categories: 
image-based \cite{li2018fast,zhu2018part,zhu20202d}, sequence-based \cite{wu2018sketchsegnet,qi2019sketchsegnet+, li2019toward, Bandyopadhyay_2024_CVPR1}, and graph-based \cite{yang2021sketchgnn,wang2020multi,zheng2023sketch} approaches, each with its own advantages and limitations. Specifically, image-based methods~\cite{li2018fast,zhu2018part,zhu20202d} use raster images as input, leveraging absolute coordinates to capture stroke proximity. However, they often struggle to learn structural information and sequential relationships among strokes effectively. Graph-based methods~\cite{yang2021sketchgnn,wang2020multi,zheng2023sketch} face challenges in capturing detailed structural information within strokes. Sequence-based methods~\cite{wu2018sketchsegnet,qi2019sketchsegnet+, li2019toward} utilize point sequences with relative coordinates, which enhances the encoding of individual stroke structures but makes it difficult to represent inter-stroke proximity and spatial relationships. 
\revi{ContextSeg \cite{wang2024contextseg} proposed the concept of the single stroke entity for sketch semantic segmentation. The advantages of this representation are demonstrated in their paper by comparing it with fancy Transformer \cite{ribeiro2020sketchformer} or common RNN \cite{li2019toward,ha2017neural} based methods.
SketchXAI \cite{qu2023sketchxai} decoupled a single stroke into its shape, location, and order embeddings, and the concatenated stroke embedding was used to train a sketch classifier to study the explainability of human sketches.
We exploit the same concept of treating a single stroke as the atomic unit, and decouple its shape with its location. Additionally and importantly, we train a vector quantized representation to build the discrete and compact code space dedicated to our generation task. 
}

\begin{figure*}[!t]
    \centering
    \includegraphics[width=1\textwidth]{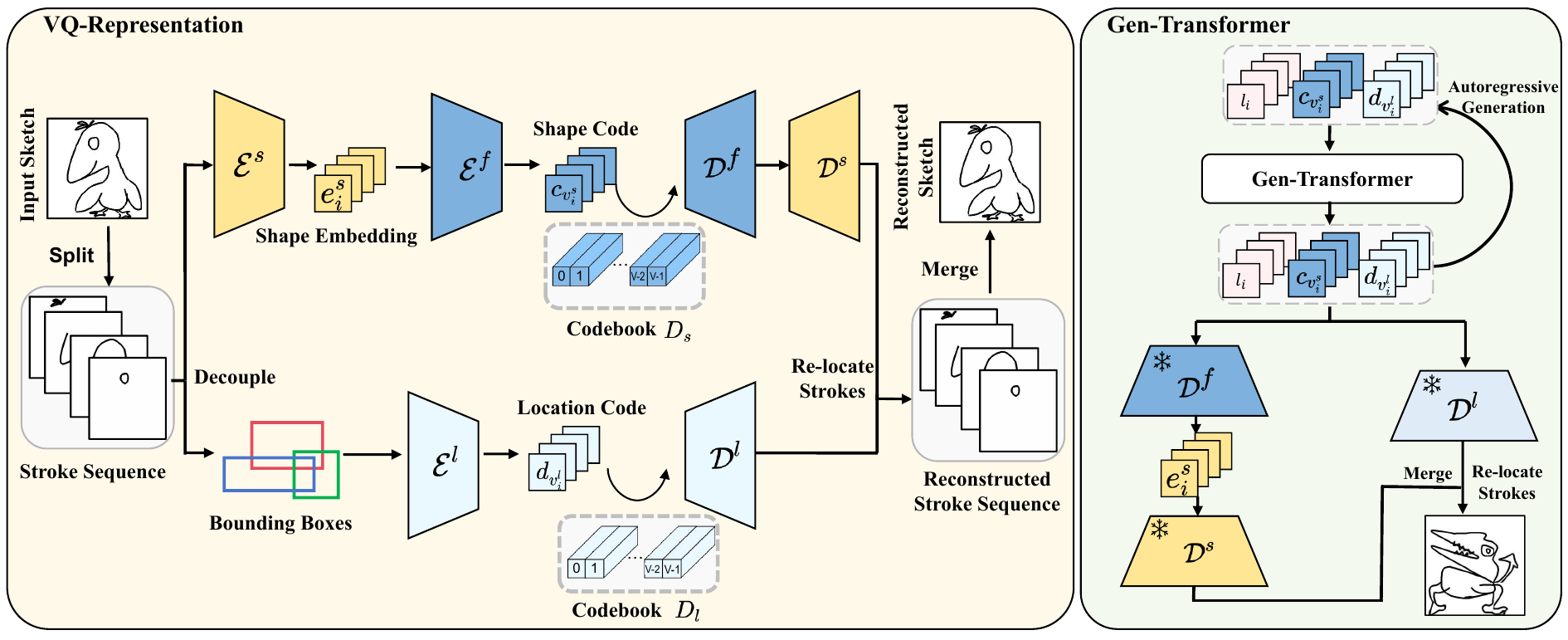}
    \vspace{-7mm}
    \caption{\textbf{Overview of \name}. Given an input sketch, it is first divided into a sequence of strokes. In the first stage, we begin by decoupling the shape and location information of each stroke and obtain their discrete representations (Sec.~\ref{subsec:discretized_latent_space}). In the second stage, we use a decoder-only Gen-Transformer to predict \( \left \{ \left( \bm{I}_i, b_i, l_i \right) \right \} \) in an autoregressive manner (Sec.~\ref{subsec:autoregressive_generation}). 
    }
    \vspace{-4mm}
    \label{pic:pipeline}
\end{figure*}

\parag{Vector-Quantized VAEs.}
Learning representations with discrete features has been the focus of many previous works \cite{hinton2006reducing, vincent2010stacked, chen2016infogan}, as such representations are a natural choice for complex reasoning, planning, and predictive learning. Although using discrete latent variables in deep learning has been difficult, effective autoregressive models have been created to capture distributions over discrete variables.
For instance, StrokeNUWA \cite{tang2024strokenuwa} utilizes VQ-VAE \cite{van2017neural} to achieve a better representation of vector graphics, then generates vector graphics through large language models. 
Similarly, we propose to learn a VQ representation for rasterized strokes, expecting an improved generation capability powered by the compact representation.

\parag{Creative Sketch Generation.} 
Unlike traditional sketch generation tasks \cite{zheng2018strokenet,zhou2018learning}, creative sketch generation \cite{ge2020creative} places greater emphasis on producing more imaginative depictions of familiar visual concepts, rather than generating conventional and mundane forms of these concepts. 
To achieve this goal, DoodlerGAN  \cite{ge2020creative} presents a component-based GAN architecture to incrementally generate each part of a creative sketch image. This approach requires training one GAN model individually for each body part in a supervised way, leading to significant computational overhead. 
DoodleFormer \cite{bhunia2022doodleformer} employs an attention-based architecture for two-stage sketch generation. 
However, this approach tends to focus only on the broad, rough relationships between different parts of the sketch, which can lead to unnatural mixing of strokes from different regions.

\section{Method}

\parag{Sketch and strokes.}
A standard sketch $\bm{S}$ in our context is composed of a sequence of strokes $\{\bm{s}_i\}_{i=1}^N$ and their corresponding \emph{stroke labels} $\{l_i\}_{i=1}^N$. 
\revi{The order of each stroke in the sequence corresponds to their default drawing order.}
Each stroke is represented as a rasterized image of size $256 \times 256$, and each label $l_i$ is a one-hot vector in $\mathbb{R}^C$.
In order to decouple the shape and position of a single stroke, we further process it. 
Formally, for each rasterized stroke image, we calculate its axis-aligned bounding box and record its center coordinate $(x_i, y_i)$. The bounding box size (\ie, the height and width) as well as the center coordinate fully represent the \emph{stroke position} $b_i=(w_i/2, h_i/2, x_i, y_i)$.
By factoring out the position (\ie, translating the stroke bounding box to align with the sketch image center), the \emph{stroke shape} itself is presented as the translated rasterized image $\bm{I}_i$ \revi{without changing its scale}.
Considering the triplet of shape, position, and semantic of a single stroke, the new format to represent a sketch is $\bm{S}= \{\bm{s}_i\}_{i=1}^N = \{(\bm{I}_i, b_i, l_i)\}_{i=1}^N$. 

\parag{Sketch generation.} 
Our goal in this paper is to generate a sequence of strokes in the format of the triplet and assemble them based on their shape and position to form the resulting sketch (see \cref{pic:teaser}). 
Figure \ref{pic:pipeline} displays an overview of our novel sketch generation method, which is comprised of two main components. Firstly, a decoupled representation learning module encodes a stroke into its vector-quantized representation (\cref{subsec:discretized_latent_space}) that is highly compact and shape-aware. Secondly, a bespoke generator based on the Transformer \cite{vaswani2017attention} is designed to generate the stroke label, shape, and location sequentially, achieving a fine-grained control with high-fidelity resulting sketches. In the following sections, we elaborate on the details.

\subsection{Vector-quantized Stroke Representation}
\label{subsec:discretized_latent_space}

Our triplet stroke formula enables separate control of stroke key properties in the generation process, how to properly encode the triplet to fit generation is the next step. Motivated by the vector quantization process \cite{van2017neural}, we aim to obtain a similar compact stroke representation in the context of the whole sketch, such that the sketch generator can efficiently and effectively sample new sketches in a compressed stroke space.
Besides, we expect a decoupled representation based on our triplet formulation to obtain high shape and semantic awareness. 
To this end, we first convert each stroke into a latent embedding and further compress them in the latent space. 

\parag{Stroke latent embedding.} 
Given the stroke image $\bm{I}_i$, we obtain the stroke latent embedding via an autoencoder, where both the encoder $\mathcal{E}^s$ and the decoder $\mathcal{D}^s$ are built with a few 2D CNNs (see supplementary for detailed network configuration). After the last layer of the encoder, we flatten its bottleneck feature to obtain the stroke latent embedding $\bm{e}_i^s = \texttt{flat}\left(\mathcal{E}^s(\bm{I}_i)\right)$. The autoencoder is trained using the reconstruction loss:
\begin{equation}
    \mathcal{L}_\text{recons} = \left\| \bm{I}_i - \mathcal{D}^s(\mathcal{E}^s(\bm{I}_i)) \right\|^{2}.
\end{equation}
To better capture the stroke shape information, we follow \cite{wang2024contextseg, liu2018intriguing, bandyopadhyay2024sketchinr} to add the CoordConv layer to the network and the distance field supervision in the training process.

\parag{Tokenizing strokes.}
To achieve fine-grained control, we build separate discretized spaces for the shape and location, respectively.
The reason to build a space for the shape location is that it is highly related to semantics, \eg, the heads of birds appear at a similar position with a similar size.

For the stroke shape space, as shown in \cref{pic:pipeline} left, we build a codebook $\bm{D}_s$. Specifically, given a sketch $\bm{S}$, its corresponding stroke shape latent embeddings $\{\bm{e}^s_i\}_{n=1}^N$ are fed into the encoder $\mathcal{E}^f$ to obtain the feature sequence $\{\bm{z}^s_i\}_{i=1}^N$. We then compress it via vector quantization, \ie, clamping each feature into its nearest code in $\bm{D}_s$ with $V$ codes $\{\bm{c}_i\}_{i=1}^V$, and we record the indices of these features:
\begin{equation}
v^s_{i} =\operatorname{argmin}_{j \in[0, V)}\left\|\bm{z}^s_i-\bm{c}_{j}\right\|.
\end{equation}
The feature sequence $\{\bm{c}_{v^s_i}\}^N$ is further sent to the decoder $\mathcal{D}^f$ to reconstruct the input latent embeddings. Both $\mathcal{E}^f$ and $\mathcal{D}^f$ are composed of several 1D CNN layers, see the supplementary for the detail structure. 

To train the vector quantization network, we follow \cite{van2017neural} to exploit  the codebook, commitment, and reconstruction losses, defined as:
\begin{equation}
    \begin{aligned}
            \mathcal{L}_\text{VQ}=\frac{1}{N} \sum_{i=1}^{N} \alpha (\|\bm{z}^s_i&-\operatorname{sg}[\bm{c}_{v^s_i}] \|_{2}^{2} + \| \operatorname{sg}[\bm{z}^s_i]-\bm{c}_{v^s_i} \|_{2}^{2}) \\
    &+ \|\bm{z}^s_i-\mathcal{D}^f(\bm{c}_{v^s_i})\|_{2}^{2},
    \end{aligned}
\end{equation}
where $\alpha$ is a weight, and $\operatorname{sg}$ is the stopgradient operator.

For the stroke location codebook learning, we use the same process as the shape codebook learning. The stroke position $b_i=(w_i/2, h_i/2, x_i, y_i)$ is reconstructed by a pair of encoder ($\mathcal{E}^l$) and decoder ($\mathcal{D}^l$) to build the codebook $\bm{D}_l$ with V codes $\{\bm{d}_i\}_{i=1}^V$ as well.  See supplementary for detailed network structures.

Till now, given a stroke $\bm{s}_i=(\bm{I}_i, b_i, l_i)$, we have the compact representation $\left(v^s_i, v^l_i, l_i\right)$, corresponding to the specific code features $\left(\bm{c}_{v^s_i}, \bm{d}_{v^l_i}, l_i\right)$.

\subsection{Autoregressive Generation}
\label{subsec:autoregressive_generation}

Having the VQ stroke representation, our goal in this module is to estimate a distribution over the sketch $\bm{S}$ so as to generate new instances. Since $S$ is composed of a sequence of stroke shapes, positions, and labels, we split the generation process into two steps, \ie, 1) generate the stroke label, and 2) generate the stroke shape and position conditioned on the stroke label. By applying the chain rule, we formulate it as:
\begin{equation}
\label{eq:sampling}
        p(\bm{S}) =  \prod_{i=1}^{N} p(\bm{I}_i, b_i, l_i) 
        = \prod_{i=1}^{N} p(v^s_i, v^l_i \mid l_i) \cdot p(l_i).
\end{equation}

We implement \cref{eq:sampling} with two cascaded \emph{Transformer decoders}, parameterized with $\theta_1$ and $\theta_2$, respectively. The first decoder samples a label for the next stroke, while the second decoder generates the shape and position conditioned on the label, which can be further formulated as:
\begin{equation}
    \begin{aligned}
        p(l_i) &= p(l_i \mid l_{<i}, v^s_{<i}, v^l_{<i};\theta_1), \\
        p(v^s_i, v^l_i \mid l_i) &= p(v^s_i, v^l_i \mid l_{\leq i}, v^s_{<i}, v^l_{<i};\theta_2).
    \end{aligned}
\end{equation}

Note that, we will use the codes in $\bm{D}_s$ and $\bm{D}_l$ in the Transformer calculation, instead of the original rasterized image and the quaternion for the stroke shape and position.

\begin{figure}[!tb]
    \centering
    \includegraphics[width=1.0\linewidth]{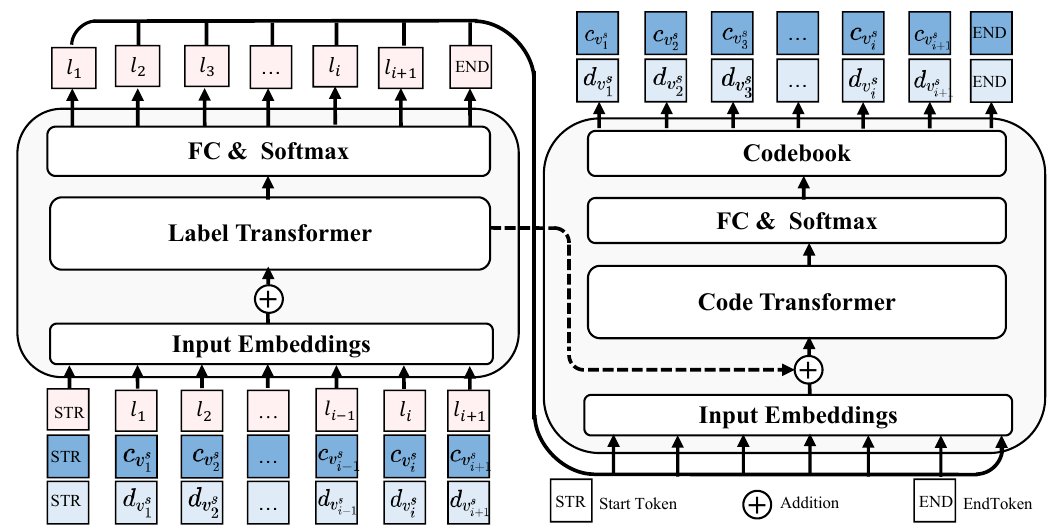}
    \vspace{-6mm}
    \caption{Network architecture of our generator. 
    }
    \vspace{-5mm}
    \label{pic:transformer_architectures}
\end{figure}

\parag{Network Architectures.} 
An overview of our cascaded decoders is shown in \cref{pic:transformer_architectures}, where the first label Transformer ($\mathcal{T}^l$) predicts a stroke label each time, while the second code Transformer ($\mathcal{T}^c$) takes as input the predicted label generating both the stroke shape and position codes simultaneously.

Concretely, the label Transformer $\mathcal{T}^l$ takes as input the triplet $\left(\bm{c}_{v^s_i}, \bm{d}_{v^l_i}, l_i\right)$. Each element goes through its own embedding layer to get a feature, \ie, $\texttt{Emb}(\bm{c}_{v^s_i}) \in \mathbb{R}^{512}$, $\texttt{Emb}(\bm{d}_{v^l_i}) \in \mathbb{R}^{512}$, and $\texttt{Emb}(l_i) \in \mathbb{R}^{512}$. The three features are simply added along the feature dimension to obtain the input feature for the attention blocks, which outputs a fused feature $\mathcal{F}_{\text{fuse}}$ after the attention calculation. $\mathcal{F}_{\text{fuse}}$ is then fed into a fully connected layer with the softmax activation to predict the one-hot probabilistic vector, which is translated into a one-hot class label $l_{i+1}\in \mathbb{R}^C$.

The code Transformer $\mathcal{T}^c$ accepts two inputs. The first one is $l_{i+1}$, which goes through an embedding layer to produce a conditional feature in $\mathbb{R}^{512}$. The second input is $\mathcal{F}_{\text{fuse}}$, which is also added with the conditional feature along the feature dimension to get the input feature for the attention block. Intuitively, the first input provides the direct semantic guidance of the current stroke, while the second input provides the shape and positional information of previous strokes. The attention blocks are followed by two separate branches, each of which outputs a one-hot probabilistic vector of dimension $\mathbb{R}^V$, indicating the indices of the shape and position codebooks. The corresponding code entries are fetched (\ie, $\bm{c}_{v^s_{i+1}}$ and $\bm{d}_{v^l_{i+1}}$) and serve as the input to the label Transformer for the next stroke generation.

\parag{Losses.} To train the network, we minimize the negative log-likelihood loss, defined as:
\begin{equation}
\mathcal{L}_{\text{gen}}=-\log p(\bm{S}), %
\end{equation}
where we supervise the training with ground truth categorical labels (\ie, the shape label vector, shape code index vector, and the position code index vector).

\parag{Training and testing strategy.} 
\revi{We have three training stages arranged sequentially for stroke latent embedding, codebooks, and the generator. \emph{Detailed steps, hyperparameter settings, and implementation details can be found in the supplementary}.}
For the testing time generation, starting from a special \emph{STR} token, we generate novel strokes in an autoregressive manner, where the codebook entry is either decoded as the position quaternion or the stroke latent code, which is further converted into the corresponding rasterized image. The stroke triplet is finally assembled together to produce the resulting sketch (see \cref{pic:pipeline}, left).

\begin{figure*}[!t]
	\centering
    \includegraphics[width=1\textwidth]{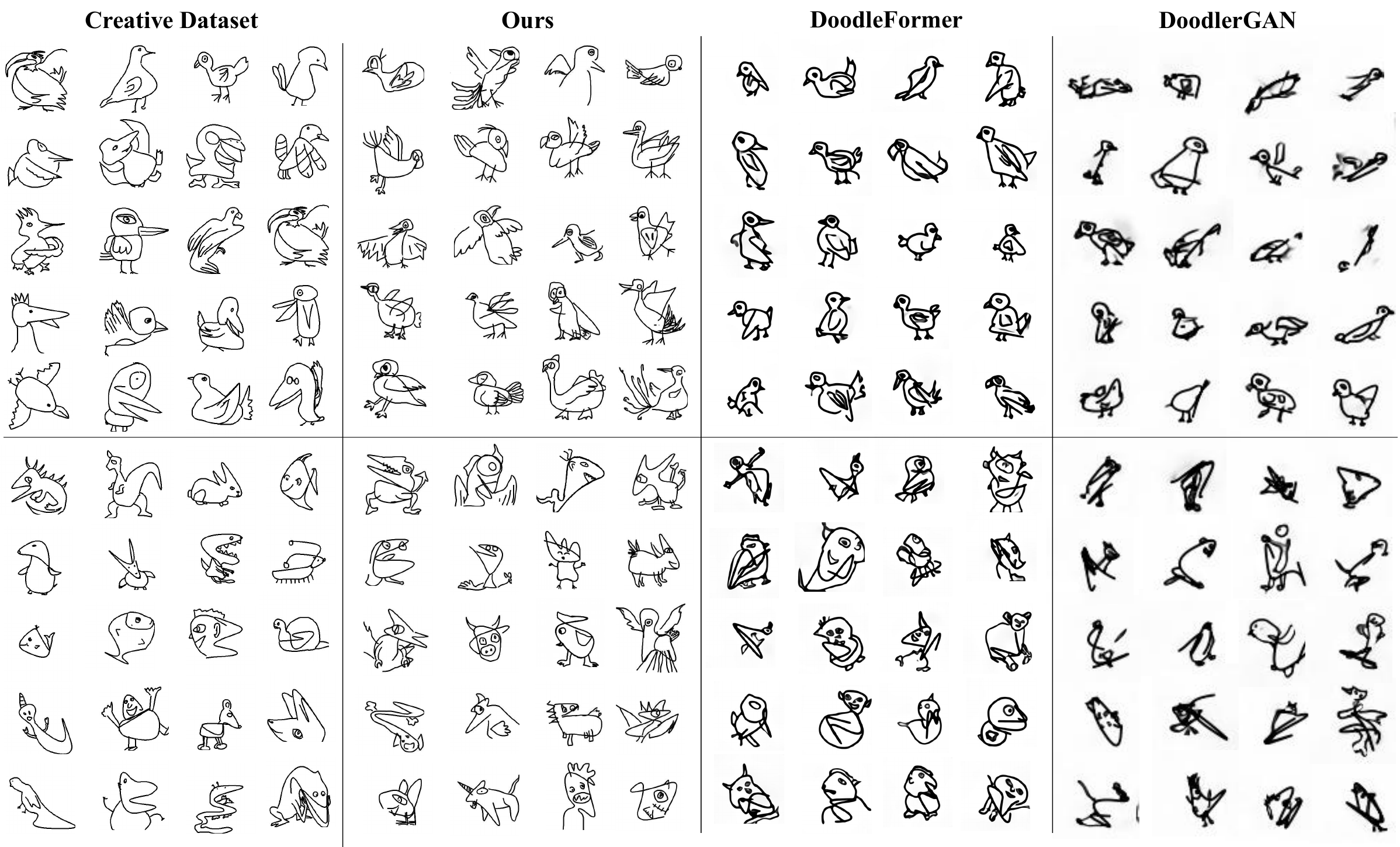}
    \vspace{-7mm}
	\caption{Visual comparison against competitors on both the Creative Birds and Creative Creatures datasets.}
    \vspace{-4mm}
	\label{pic:main_result2}
\end{figure*}

\section{Experiments}
\label{sec:exps}

\parag{Datasets.} Following previous work~\cite{ge2020creative, bhunia2022doodleformer}, we use the CreativeSketch~\cite{ge2020creative} dataset in this paper. 
Specifically, it consists of two categories of sketches — Creative Birds (CB) and Creative Creatures (CC) — containing 8,067 and 9,097 sketches, respectively, along with part annotations. 
All the sketches have been processed into 256 × 256 images. 
For the stroke latent embedding learning, we use strokes from both datasets to obtain a powerful pair of $\mathcal{E}^s$ and $\mathcal{D}^s$. Other than this autoencoder, we have trained other networks separately on CB and CC datasets.
See supplementary for details about pre-processing and data augmentation.

\parag{Evaluation metrics.} Following previous work~\cite{ge2020creative, bhunia2022doodleformer}, we evaluate the model's performance using an Inception model ~\cite{szegedy2016rethinking} trained on the QuickDraw3.8M dataset~\cite{xu2022deep}. We use four metrics in this paper, defined as follows:
\begin{itemize}
    \item \textbf{Fréchet Inception distance (FID)~\cite{heusel2017gans}} measures the similarity between two image sets in the above Inception feature space. 
    \item \textbf{Generation diversity (GD)~\cite{cao2019ai}} measures the average pairwise Euclidean distance between the Inception features of two subsets, reflecting the diversity of the data item in the image set.
    \item \textbf{Characteristic score (CS)} measures how frequently a generated sketch is classified as Creative Bird or Creative Creature by the Inception model.
    \item \textbf{Semantic diversity score (SDS)} quantifies sketch diversity based on various creature categories they represent.
\end{itemize}

\begin{table}[!tb]
\centering
\caption{Statistical comparison against existing methods on Creative Birds and Creative Creatures datasets.}
\vspace{-2mm}
\label{tab:main_results}
\renewcommand{\arraystretch}{1.2}
\resizebox{1.\linewidth }{!}{
\begin{tabular}{c|ccc|cccc}
\toprule[0.4mm]
\multirow{2}{*}{Methods} & \multicolumn{3}{c|}{Creative Birds}& \multicolumn{4}{c}{Creative Creatures}\\
\cline{2-8}

& FID(↓) & GD(↑) & CS(↑) & FID(↓) & GD(↑) & CS(↑) & SDS(↑) \\

\hline
Training Data & - & 19.40 & 0.45 & - & 18.06 & 0.60 & 1.91 \\
\hline

Sketchknitter~\cite{wang2023sketchknitter} & 74.42 & 14.23 & 0.14 & 64.34 & 12.34 & 0.42 & 1.32\\

DoodlerGAN~\cite{ge2020creative}& 39.95 & 16.33  & \textbf{0.69}  & 43.94 & 14.57 & 0.55 & 1.45 \\

DoodleFormer~\cite{bhunia2022doodleformer} & 17.48 & 17.83 &  0.57 & 20.43 & 16.23 &0.53 & 1.68\\

\hline

Ours & \textbf{15.78} &\textbf{18.92} & 0.53& \textbf{17.61} &\textbf{17.42} & \textbf{0.57}& \textbf{1.86} \\
\bottomrule[0.4mm]
\end{tabular}
}
\vspace{-4mm}
\end{table}

\subsection{Comparison}

We compare our approach with three competitors, \ie, Doodleformer~\cite{bhunia2022doodleformer}, DoodlerGAN~\cite{ge2020creative}, Sketchknitter~\cite{wang2023sketchknitter}. Both Doodleformer and DoodlerGAN generate sketches at the pixel level, while Sketchknitter generates sketches at the stroke point sequence level. All three methods are trained using their default parameters on our dataset.

The quantitative comparison is presented in \cref{tab:main_results}, where our method significantly outperforms others. 
Specifically, on the Creative Birds dataset, we obtained a notable 1.7 decrease in FID and gained a remarkable 1.09 increase in terms of GD, achieving the best performance. 
This indicates that our model can generate sketches that are closer to the original dataset, as well as more diverse at the same time. 
For the CS metric, while our approach does not achieve the highest score, it surpasses the dataset and comes closest to matching the dataset's CS metric. In contrast, DoodlerGAN and DoodleFormer attain higher scores, whereas SketchKnitter's outputs are seldom recognized as birds. For this specific metric, the higher value only means it is easier to be recognized by the Inception model, which may stem from the simplicity of the generation. The dataset's value sets a good reference.
Similarly, on the Creative Creatures dataset, we achieved the best performance across all four metrics. 
Specifically, we achieved a 2.82 decrease in FID and a 1.19 improvement in GD compared to the best other methods. This indicates that our generated sketches are closer to the more complex sketches in the training data and exhibit greater diversity among themselves. For the CS and SDS metrics, we obtained improvements of 0.04 and 0.18, respectively. This indicates that our generated sketches are semantically closer to the categories in the training data and also suggests that our model exhibits better semantic diversity. 
A corresponding visual comparison is shown in \cref{pic:main_result2}, where the artifacts like the broken and misaligned strokes from DoodlerGAN, and the blurry local regions from DoodleFormer, are clearly presented. In contrast, our results are of high quality with complex appearance and diverse shapes. Note that, as shown in \cref{pic:teaser}, SketchKnitter generates poor results on both datasets, we did not include its results in the visual comparison.

\begin{table}[!t]
\centering
\caption{The statistical comparison of our ablation study. The method name of two numbers represents the codebook hyper-parameter setting, where the first is the code size, and the second is the feature dimension.} %
\vspace{-2mm}
\label{tab:ablation_results}
\renewcommand{\arraystretch}{1.2}
\resizebox{1.\linewidth }{!}{
\begin{tabular}{c|ccc|cccc}
\toprule[0.4mm]
\multirow{2}{*}{Methods} & \multicolumn{3}{c|}{Creative Birds}& \multicolumn{4}{c}{Creative Creatures}\\
\cline{2-8}

& FID(↓) & GD(↑) & CS(↑) & FID(↓) & GD(↑) & CS(↑) & SDS(↑) \\

\hline
Training Data & - & 19.40 & 0.45 & - & 18.06 & 0.60 & 1.91 \\
\hline

2048$\times$512  & 26.23 & 15.34 & 0.43 & 43.21 & 14.51 & 0.48 & 1.52\\

4096$\times$512 & 16.92 & 18.27  & 0.50  & 18.44 & 16.14 & 0.54 & 1.63 \\

4096$\times$1024 & 16.67 & 18.15 &  0.51 & 18.12 & 16.43 &0.55 & 1.69\\

\hline
w/o $\mathcal{T}^l$ & 16.51 & 18.35 &  0.51 & 19.21 & 16.14 &0.55 & 1.74\\
w/o VQ & 48.53 & 13.34 &  0.46 & 54.56 & 14.02 & 0.45 & 1.36\\

w/o Decouple & 17.14 & 18.12 &  0.57 & 19.42 & 16.42 & 0.56
 & 1.79 \\
\hline

Ours($8192\times512$)& \textbf{15.78} &\textbf{18.92} & \textbf{0.53}& \textbf{17.61} &\textbf{17.42} & \textbf{0.57}& \textbf{1.86} \\
\bottomrule[0.4mm]
\end{tabular}
}
\vspace{-5mm}
\end{table}

\begin{figure*}[!htb]
    \centering
    \includegraphics[width=0.9\textwidth]{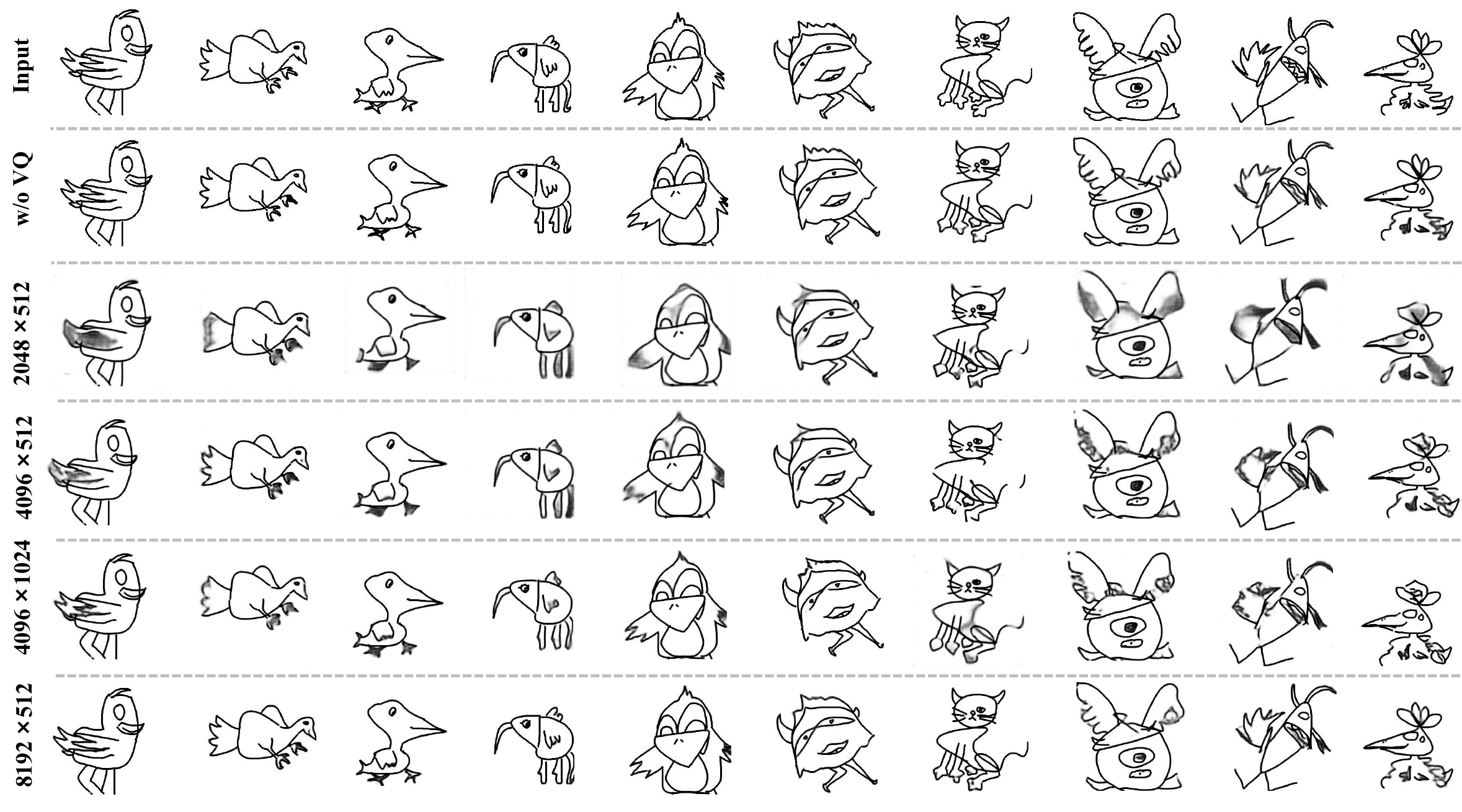}
    \vspace{-4mm}
    \caption{Sketch reconstruction results of our ablation study on codebook hyper-parameter setting. The two numbers represent the code size and the feature dimension, respectively. 
    }
    \label{pic:ablation}
    \vspace{-4mm}
\end{figure*}

\subsection{Ablation Study}
\label{subsec:abl_study}
To validate the effectiveness of our core technical components, we conducted experiments by removing them. 
Specifically, we form the following alternative solutions: 
\begin{itemize}
    \item w/o VQ: we remove all the codebook learning-related procedures, and use the continuous stroke latent embeddings to train our generator.
    \item w/o Decouple: we do not decouple the shape and position of the stroke, and only train a single stroke codebook. The Transformer decoders take fewer inputs and the second decoder outputs only one index vector accordingly. 
    \item w/o $\mathcal{T}^l$: we simply discard the label information for each stroke, and remove the label Transformer accordingly.
\end{itemize}

We train the three alternatives from scratch using our datasets until convergence and statistical results are shown in \cref{tab:ablation_results}. 
Without the VQ representation, the generator samples new sketches from a larger continuous space, instead of a compressed discretized space, giving the worst performance in this paper. 
The stroke decoupling matters the generation significantly (\eg, 15.78 $vs.$ 17.14 and 17.61 $vs.$ 19.42 of the FID). By decoupling, the shape code can concentrate only on the shape variation, while the position code can better capture the inherent approximation information between strokes, which further facilitates the generation.
By removing the label prediction, we see a slight performance degradation, giving the second-best performance in \cref{tab:ablation_results}. The labels are not very critical, but indeed provide semantic-related information to complement the positional information (\eg, the heads appear at similar positions with similar sizes). 
\revi{On the other hand, it demonstrates the great feasibility of applying our method to other datasets without label information, proving our flexibility. See Sec. \ref{sec:discuss} for such an experiment.}

\parag{Codebook hyper-parameter setting.}
There are two core hyper-parameters for a codebook, \ie, the code size and the feature dimension. 
\revi{In the following, we have conducted experiments on the shape codebook to find the proper configuration. Similar experiments on the location codebook are presented in the supplementary.}
Specifically, we set the code size to be either 2,048, 4,096, or 8,192, and set the feature dimension to be either 512 or 1,024.
We examine the effects of the parameter setting from both the sketch reconstruction and generation perspectives.

\begin{figure}[!tb]
    \centering
    \includegraphics[width=1.0\linewidth]{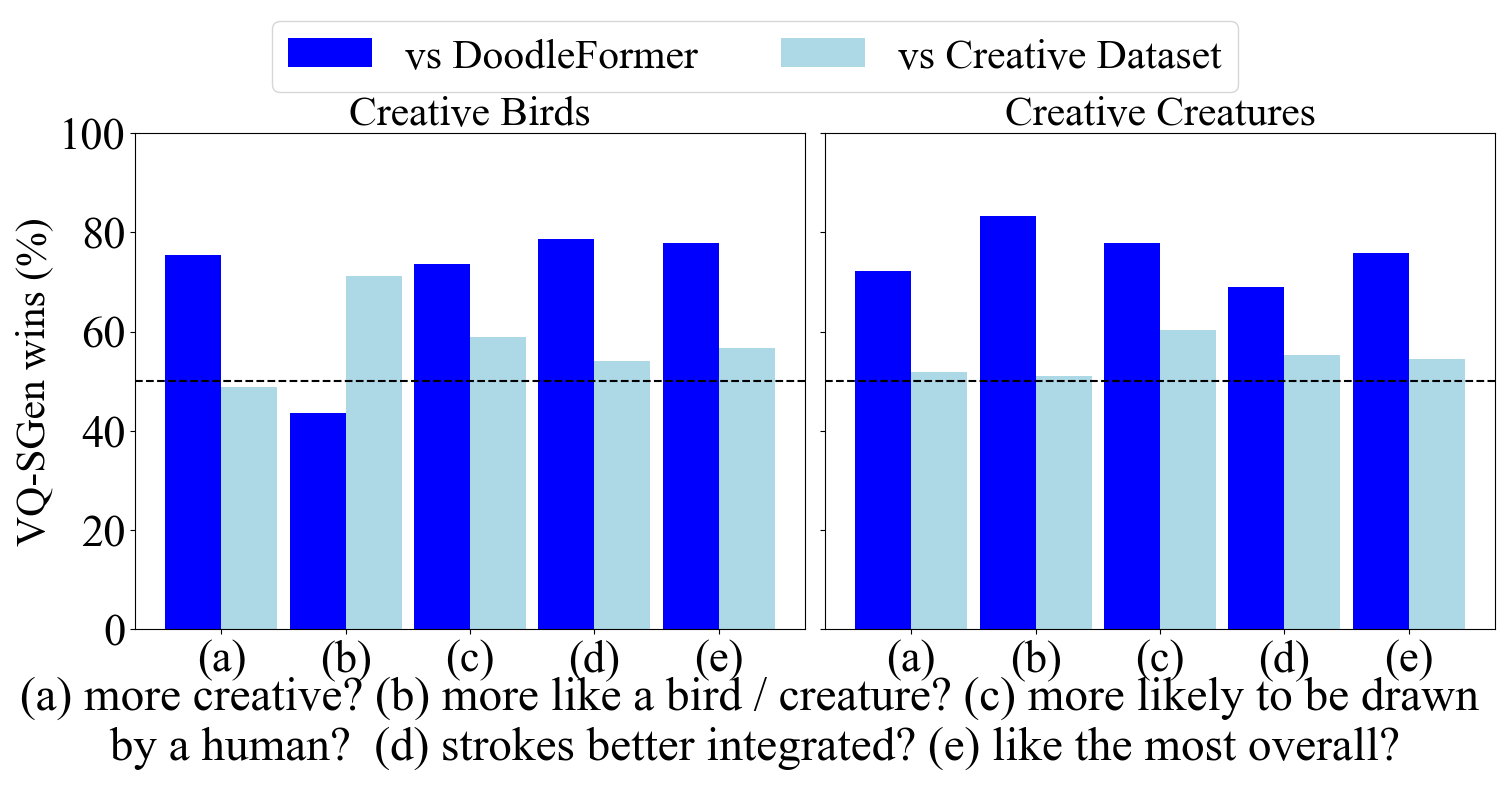}
    \vspace{-7mm}
    \caption{User study results based on the five questions (a-e). Higher scores indicate a stronger preference for our method over the other approaches (\ie, DoodleFormer or Creative datasets).}
    \vspace{-6mm}
    \label{pic:user_study}
\end{figure}

Figure \ref{pic:ablation} presents a visual comparison of sketch reconstruction quality. By increasing either the code size or the feature dimension, we observe improved fidelity with less blurry and broken strokes. Note that the row named \emph{w/o VQ} shows the stroke latent space reconstruction performance as in \cite{wang2024contextseg}, which along with the input, provides a good performance reference. We chose the last configuration in this paper, as it gives the best visual reconstruction. However, it still struggles to represent strokes with significant variation, such as the ear of the creature in the third-to-last column.
We further investigate the generation performance of different codebook settings. Quantitative results are listed in \cref{tab:ablation_results}, where we observe improvement when increasing either the size of the code or the feature dimension. Our choice obtains the best result given the limited combination.

\begin{figure*}[!htb]
    \centering
    \includegraphics[width=0.99\textwidth]{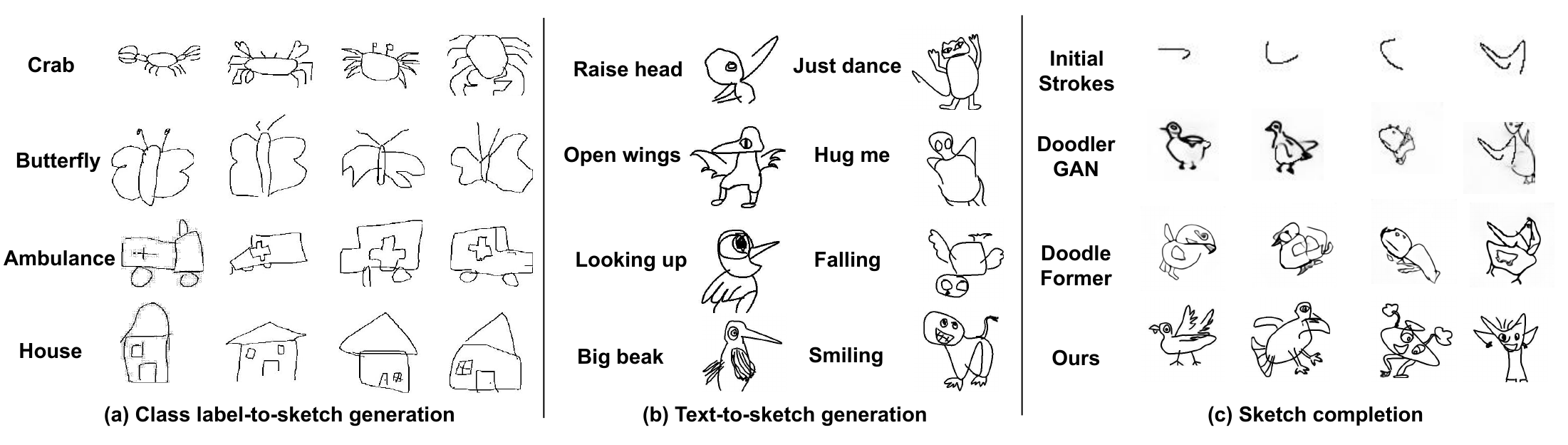}
    \vspace{-4mm}
    \caption{Applications of our approach. (a) Given the class label, our method can produce corresponding sketches with rich variations. (b) Our method supports text-conditioned generation, the resulting sketches explain the input text vividly. (c) Given the initial stroke, our method can complete the whole sketch, which is favorable against competitors.
    }
    \vspace{-3mm}
    \label{pic:applications}
\end{figure*}

\subsection{User Study} 
Following \cite{bhunia2022doodleformer,ge2020creative}, we evaluate the performance of our method through user studies. 
We invited 50 participants to the test, where they were shown multiple pairs of images. Each pair consists of a sketch generated by our model and another sketch from either DoodleFormer or the datasets. For each pair, participants were asked to answer 5 single-choice questions: which one (a) is more creative? (b) looks more like a bird/creature? (c) is more likely to be drawn by a human? (d) the strokes better integrated? (e) they like the most overall? 
Figure \ref{pic:user_study} shows the percentage of times our method is preferred over the competing approach. It can be seen that \name performs favorably against DoodleFormer for all five questions on both datasets except for question (b) on the Creative Birds dataset, which is consistent with the CS metric in \cref{tab:main_results}. We are nearly tied with the creative datasets.

\begin{figure}[!tb]
    \centering
    \includegraphics[width=0.94\linewidth]{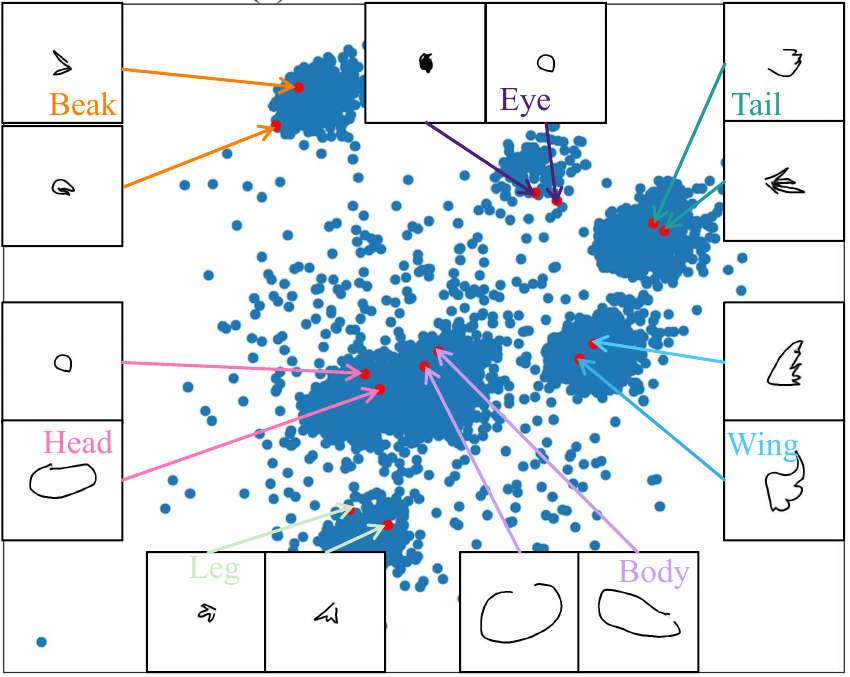}
    \vspace{-2mm}
    \caption{UMAP visualization of the shape codes in $\mathcal{D}^s$, with the overlaid strokes indicating the semantic aware clustering in the discrete code space.
    }
    \vspace{-5mm}
    \label{pic:code_vis}
\end{figure}

\section{Discussion and Application}
\label{sec:discuss}

\revi{In this section, we mainly discuss code space exploration and conditional generation applications, while leaving other discussions, \eg, limitations, in the supplementary.}

\parag{Code space exploration.} 
To better understand the properties of the discrete space, we applied the UMAP algorithm \cite{mcinnes2018umap} to project the high dimensional features of the shape codebook $\bm{D}^s$ into 2D space and visualize its distribution in \cref{pic:code_vis}. We further probe the emerged clustering by visualizing typical strokes. Not too surprised, we find that the emerged clusters correspond well to semantics. For example, the five surrounding clusters are \emph{Beak},  \emph{Eye}, \emph{Tail}, \emph{Wing}, and \emph{Leg}, while the inner big cluster composed of \emph{Head} and \emph{Body}. Although we did not exploit any semantic-related constraints during the codebook learning, it tends to implicitly group strokes with similar shape variations, thereby achieving stable and efficient code learning and space compression.
As a result, this provides an ideal foundation for the generator to sample new strokes within a semantically-aware, compressed space.

\parag{Class label-to-sketch generation.}
\revi{Other than unconditional generation, our method can generate sketches given the class label conditions. To do this, we exploit a subset of the QuickDraw dataset \cite{ha2017neural}. The subset contains 20 classes, and stroke labels are \emph{NOT} available in this dataset.
Our adaptation is pretty simple and straightforward - i) removing the label Transformer (\ie, the variant of w/o $\mathcal{T}^l$ in Sec. \ref{subsec:abl_study}), and ii) replacing the special \emph{STR} token with the class index token (\eg, the sinusoidal encoding). We only re-train the generator on the QuickDraw subset until convergence. At inference time, given a class label, our method successfully produces high-quality results, as shown in Fig. \ref{pic:applications}(a). Note that, the experiment also demonstrates our flexibility and superior performance when working on datasets without stroke label information.
}

\parag{Text-to-sketch generation.}
\revi{There are text descriptions for each sketch in the CB and CC datasets. We thus adapt our method to perform text-to-sketch generation. To this end, we first embed the text description using the pre-trained CLIP \cite{radford2021learning} model, and then replace the \emph{STR} token with the obtained text token, similarly as in the above label-to-sketch generation task.
Fig. \ref{pic:applications}(b) shows a few novel examples of our generated sketches, which conform well to input texts. Pay attention to the novel visual concept, \eg, the `Hug me' sketch, and the `smiling' sketch, which is rarely seen in the traditional sketch generation task. 
}

\parag{Sketch completion.} 
Given an initial stroke, our model can perform sketch completion, \ie, stroke conditional generation. 
We achieve this by feeding the initial stroke as our first stroke in the generation process.
\cref{pic:applications}(c) presents a visual comparison against SoTA methods on this task. As can be seen, DoodlerGAN produced the worst results, with most sketches lacking a clear structure. DoodleFormer performed slightly better, but the generated sketches are often blurry and include ghost strokes (\eg, a few tail-like strokes of the second example). With the help of our VQ representation and our generator, our results are both creative and structurally coherent.

\section{Conclusion}
In this paper, we have presented a novel approach for creative sketch generation. Our method can effectively create sketches with an appealing appearance and coherent structures. We believe that stroke VQ representations can be used in other sketch-related tasks and hope to inspire future research in sparse stroke representation learning.

\paragraph{Acknowledgment.}
The authors would like to thank Oisin Mac Aodha and Ankan Bhunia for proofreading the early draft and the valuable discussions, Haocheng Yuan and Yi Yang for helping prepare comparisons with Diffsketcher. CL was supported by a gift from Adobe.

{
    \small
    \bibliographystyle{ieeenat_fullname}
    \bibliography{main}
}

\clearpage
\appendix
\setcounter{page}{1}

\setcounter{table}{0}
\renewcommand{\thetable}{A\arabic{table}}
\setcounter{figure}{0}
\renewcommand{\thefigure}{A\arabic{figure}}

\maketitlesupplementary

In this supplemental material, we provide further details on dataset processing, technical design, codebook exploration, and additional discussions.  

\section{Data Preprocessing and Augmentation}

\begin{figure}[!htb]
    \centering
    \includegraphics[width=1.0\linewidth]{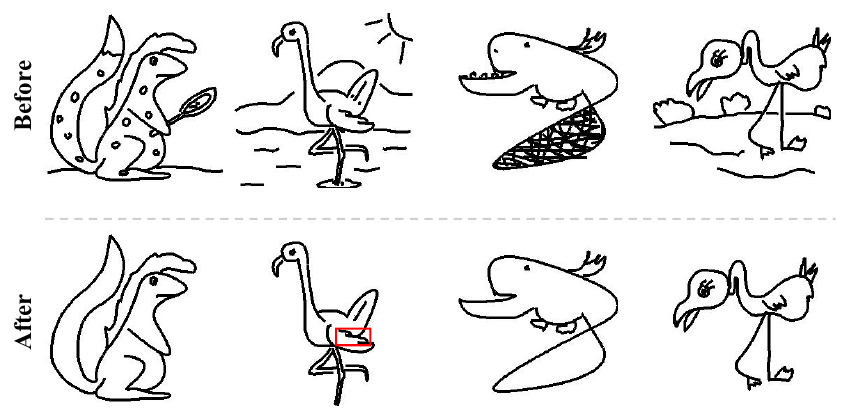}
    \vspace{-7mm}
    \caption{The demonstration of our data preprocessing.}
    \vspace{-6mm}
    \label{pic:data_preprocsssings}
\end{figure}

To enhance the model's performance and improve the quality of the training data, we adopt a series of preprocessing steps as shown in \cref{pic:data_preprocsssings}. 
Firstly,  strokes that have the 'details' part label (\eg, the sun, the ground, and the decorative dots) are moved, as these may introduce noise or unnecessary complexity. 
Secondly, excessively long strokes are eliminated to maintain a balanced distribution of stroke lengths (\eg, the over-sketched pattern in the tail of the third creature), preventing outliers from disproportionately influencing the model. 
Lastly, shorter strokes that are connected are merged into a single stroke to maintain a better stroke structure. For instance, the wing in the second image (highlighted with a red box) is composed of multiple connected strokes that are merged into a single stroke, which ensures more coherent and natural groupings within the sketches.

Besides preprocessing, we employ a comprehensive augmentation strategy at both the stroke and sketch levels. At the stroke level, transformations such as rotation, translation, or scaling are applied to one or more randomly selected strokes. These operations diversify the stroke representations and improve the model's ability to generalize to various configurations. At the sketch level, the entire sketch is subjected to global transformations, including rotations or random removal of specific strokes, simulating incomplete or imperfect inputs. 

\section{Training and Inference Details} 

\parag{Implementation details.} 
We use Adam optimizer \cite{kingma2014adam} for training both the VQ-Representation and Gen-Transformer, while the learning rates are set as $10^{-4}$ for the former and $10^{-5}$ for the latter. We use step decay for VQ-Representation with a step size equal to 10 and do not apply learning rate scheduling for Gen-Transformer.
We train our network on 2 NVIDIA V100 GPUs. The VQ-Representation network training takes around 20 hours with a batch size of 64. The Gen-Transformer is trained until convergence taking around 10 hours with a batch size of 8. 
The maximum stroke length $N$ is 20 for Creative Birds and 35 for \reviICCV{Creative Creatures} to make sure it fits all the sketches. Stroke latent embedding $\bm{e}_i^s$ is a vector in $\mathbb{R}^{256}$. The number of codes $V$ = 8192, and the code dimensions of both codebooks are $512$. Stroke category number $C=8$ for Creative Birds and 17 for Creative Creatures. The balancing weight $\alpha=0.8$.

\parag{Training Strategy.}
We have three training stages. In the first stage, we train the stroke latent embedding auto-encoder. Once it is trained, both the encoder and decoder are fixed. In the second stage, we train the vector quantization auto-encoders and the corresponding codebooks. Similarly, once the network is trained, the three components are fixed. In the last stage, we train the cascaded generation decoders together, with the help of the aforementioned networks and codebooks. Note that, we follow \cite{mihaylova2019scheduled} to address the teacher-forcing gap.

\parag{Teacher Forcing Gap.}
Teacher-forcing strategy is widely used for Transformer training. However, it introduces the exposure bias issue by always feeding the ground truth information to the network at training time but exploiting the inferior prediction at testing time.  To overcome this issue, we follow \cite{mihaylova2019scheduled} to mix the predicted stroke information with the ground truth information. The ratio of the ground truth strokes gradually decreases from 1.0 as the training progresses.

\parag{Sampling at Inference.} Starting with a special \emph{STR} token, the inference process of our \name generates a sketch by sequentially sampling each subsequent element of the stroke sequence until reaching an \emph{END} token. Each prediction step consists of two stages: (1) generating a minimal set of codes for sampling, comprising the top options whose cumulative probabilities exceed a threshold $p_n$, while ignoring other low-probability codes; (2) performing random sampling within this minimal set based on the relative probabilities of the codes. This approach ensures diversity in the predicted code sequence while reducing the risk of sampling unsuitable or irrelevant codes.

\section{Location Codebook Probing}
\begin{figure}[!htb]
    \centering
    \includegraphics[width=0.99\linewidth]{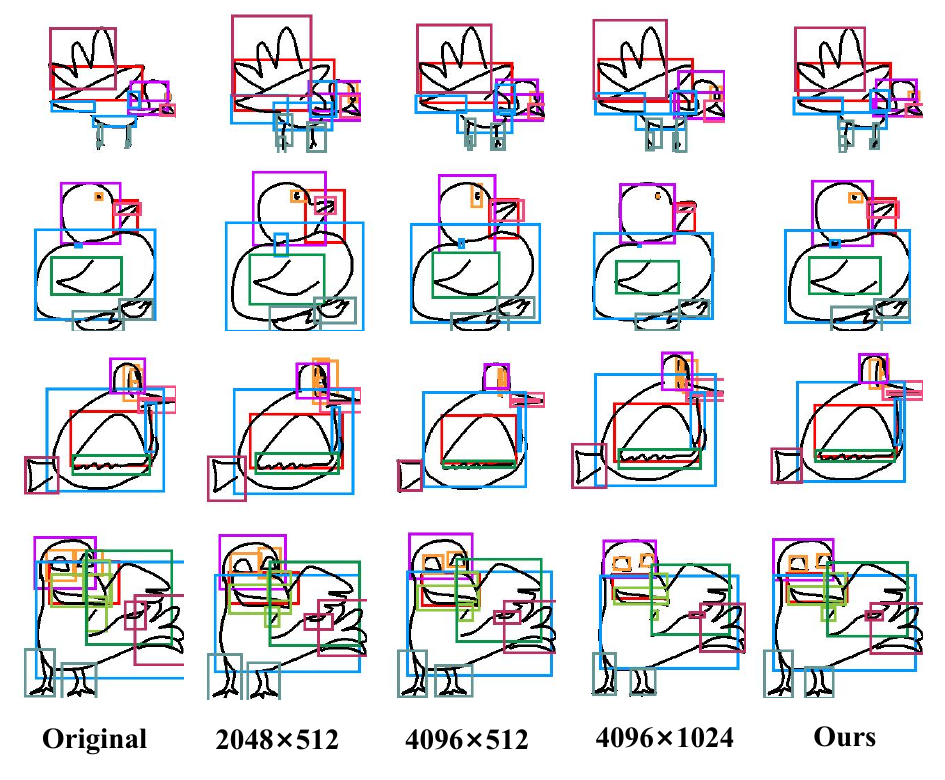}
    \vspace{-3mm}
    \caption{Visualization results of bounding box reconstruction. Note that the sketches are not generated, we overlay the input sketch to better perceive the bounding box placement. 
    }
    \label{pic:suppl_bbx}
\end{figure}

\revi{
Similar to the shape codebook, we investigated the configuration of the location codebook, in terms of the effects of its hyper-parameter setting on its representation ability and sketch generation.
For the codebook size, we choose 2,048, 4,096, or 8,192, while for the feature size, we choose 512 or 1,024.
Quantitative and qualitative experiments as shown in Fig. \ref{pic:suppl_bbx} and Tab. \ref{tab:rebuttal_bbox}, respectively.

For the codebook representation ability, we investigate it via its reconstruction. For any location triplet, we recover its bounding box representation. \cref{pic:suppl_bbx} displays the input and reconstructed bounding boxes. We overlay the input sketch for better visualization. As can be seen, varying the codebook size or the feature size indeed impacts the representation ability, while our final choice obtains the best reconstruction. 
Additionally, we employ the mean IoU (Intersection over Union) of bounding boxes as a statistical measurement in \cref{tab:rebuttal_bbox}, where the best IoU is achieved from our final choice. As for the sketch generation, we only measure the statistical metrics as in \cref{tab:rebuttal_bbox}. Not surprisingly, our choice is the best. 
Although increasing the codebook size further may potentially enhance generation quality, similar to the findings with the shape codebook, we have not yet confirmed whether a larger codebook consistently yields performance improvements. Therefore, selecting the appropriate codebook size requires balancing generation quality and computational cost and may depend on the specific requirements of the task.
A similar conclusion is drawn as in the shape codebook analysis, \ie, we did not enumerate all the configurations, and thus cannot guarantee the optimal configuration. 
}

\begin{table}[!tb]
    \centering
    \caption{Statistical comparison and ablation studies of the hyperparameter setting of the location codebook.}
    \vspace{-2mm}
    \setlength{\tabcolsep}{2pt} %
    \renewcommand{\arraystretch}{1.2}
    \resizebox{\linewidth}{!}{
    \begin{tabular}{@{}c|cccc|ccccc@{}}
        \toprule[0.4mm]
        \multirow{2}{*}{Methods} & \multicolumn{4}{c|}{Creative Birds}& \multicolumn{5}{c}{Creative Creatures}\\
        \cline{2-10}

        &IoU(↑)& FID(↓) & GD(↑) & CS(↑) &IoU(↑) & FID(↓) & GD(↑) & CS(↑) & SDS(↑)   \\

\hline
2048$\times$512& 0.912 & 17.43 & 17.54 & 0.47 & 0.901 & 20.34 & 15.75& 0.51& 1.72\\

4096$\times$512& 0.943 & 16.15  & 18.68  & 0.51 & 0.923 & 18.32 & 16.54&  0.55& 1.79\\

4096$\times$1024 & 0.954 & 16.31 &  18.53 & 0.52 & 0.934 &18.07 & 16.86& 0.56 & 1.81\\

\hline

Ours($8192\times512$)& \textbf{0.963} &\textbf{15.78} & \textbf{18.92}& \textbf{0.53} &\textbf{0.957} & \textbf{17.61}& \textbf{17.42} & \textbf{0.57 }& \textbf{1.86}\\
\bottomrule[0.4mm]
\end{tabular}
}
\vspace{-2mm}
\label{tab:rebuttal_bbox}
\end{table}

\begin{figure}[!htb]
    \centering
    \includegraphics[width=0.99\linewidth]{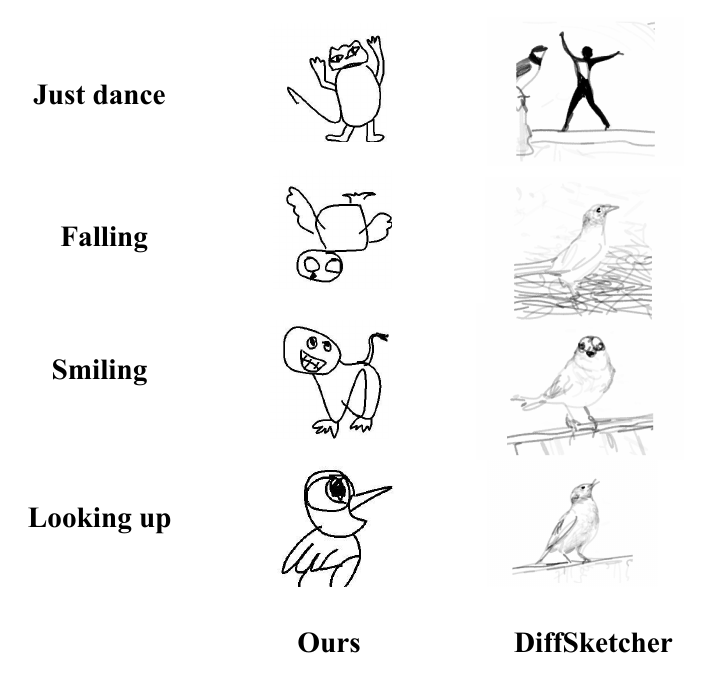}
    \vspace{-3mm}
    \caption{The visual comparison with Diffsketcher \cite{xing2023diffsketcher} on the text-to-sketch application.
    }
    \label{pic:comp_diffsketcher}
\end{figure}

\begin{figure}[!htb]
    \centering
    \includegraphics[width=0.9\linewidth]{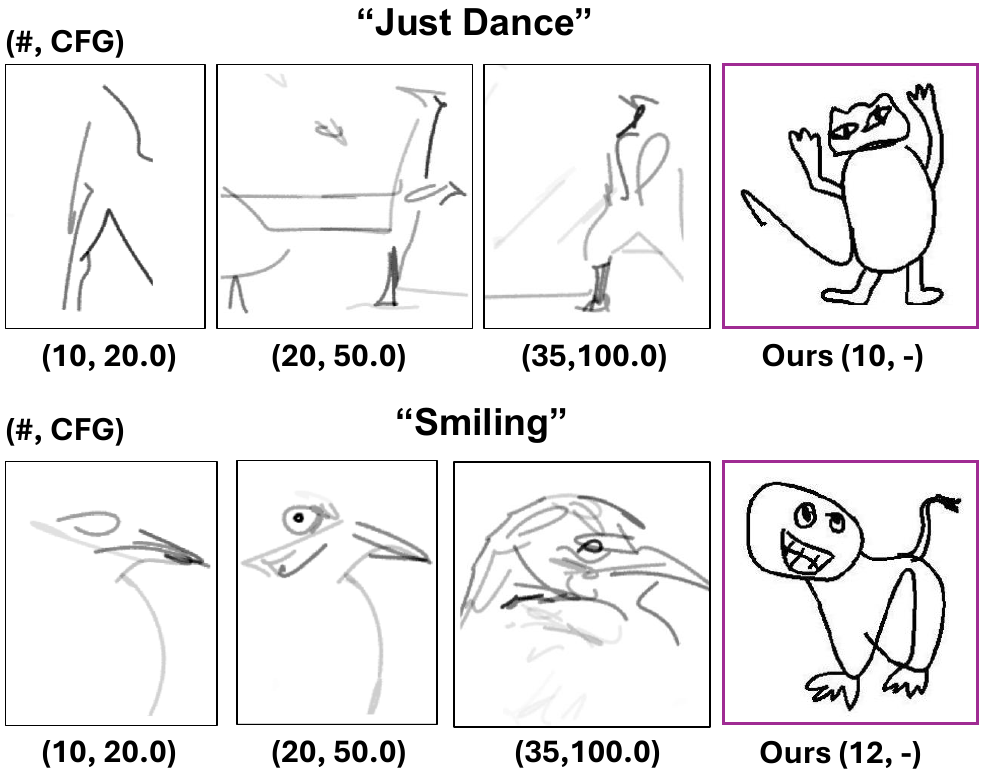}
    \vspace{-3mm}
    \caption{More DiffSketcher comparison \cite{xing2023diffsketcher} with varying  numbers of strokes and CFG weights. }
    \label{pic:comp_num_strokes}
\end{figure}

\section{Creativity Comparison with Diffsketcher }
\revi{As presented in Sec. 5 in the main paper, our method can be easily adapted to do text-to-sketch generation, which aligns not exactly the same but closer with the goal of Diffsketcher \cite{xing2023diffsketcher}. 
However, we argue that the sketches they generated possess high realism and conform to conventional imagination since they are trained on realistic images.
\cref{pic:comp_diffsketcher} displays a comparison. 
For Diffsketcher, we use a common prompt template - `A bird is [input text]', except [Just dance], for which we use `A bird is \emph{performing} [just dance]'. Different random seeds and prompt templates are tested, but their sketches do not differ too much.

As can be seen, our sketches, \eg, the smiling bird or the falling bird, are much more plausible, conforming with input text and representing more creative concepts, while their corresponding sketches cannot faithfully explain the text, limited to conventional bird sketches. Besides, their sketches usually have background strokes due to a different goal in the generation, \eg., the tree branches. Especially, the dancing person is wrongly generated by their model, since it was fooled by the word dance, and cannot link the concept of dance with a bird.
On the other hand, our method is over \emph{two magnitudes} faster (\ie, $0.86s$ vs. $153s$), making us both efficient and effective in the text-driven sketch generation task.
}

\reviICCV{In order to generate more abstract sketches comparable to ours, two extra hyperparameters can be tuned in DiffSketcher - the number of strokes and the classifier-free guidance (CFG) weight. As shown in Fig.~\ref{pic:comp_num_strokes}, we gradually increase the number of strokes as well as the CFG weight in two examples.
When the stroke count is low (10 or 20 strokes), their sketches are too abstract to be recognizable. The smiling bird with 35 strokes starts focusing on realistic head details, but it falls short of creativity.}

\section{Further discussions}
\begin{figure}[!t]
    \centering
    \includegraphics[width=0.99\linewidth]{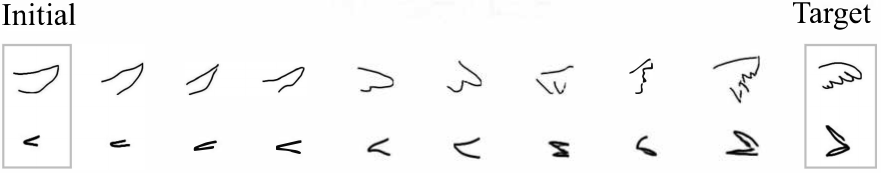}
    \vspace{-2mm}
    \caption{Two code interpolation examples, where the interpolation is executed from the initial to the target strokes (left to right). 
    }
    \label{pic:interpolation}
\end{figure}

\parag{Code space interpolation.}
To further validate the clustering effects of the discrete code space, we conduct code interpolation experiments. \cref{pic:interpolation} demonstrates two such examples. The interpolation is executed from the initial stroke to the target stroke (left to right). We first use linear $\alpha$-blending to mix the two codes, and then fetch the closest code entry in $\mathcal{D}^s$ and decode it into a stroke image. We march 10 steps to achieve the target. As can be seen, the initial and target strokes can be meaningfully interpolated in the discrete code space.

\begin{figure}[!h]
    \centering
    \includegraphics[width=0.9\linewidth]{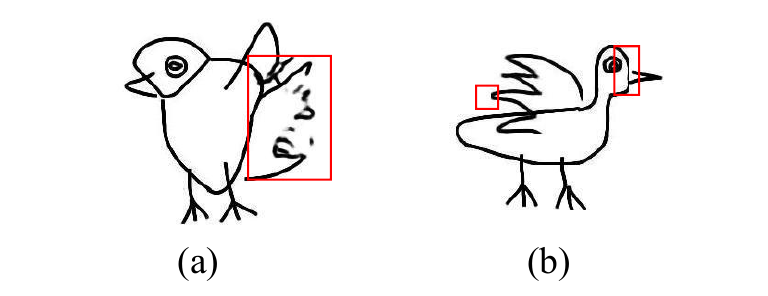}
    \vspace{-4mm}
    \caption{The demonstration of failure cases.}
    \label{pic:failure_case}
\end{figure}

\parag{Limitations.}
Our method has the following limitations: 
First, we did not extensively explore all combinations of codebook hyperparameters, so our current configuration may not be globally optimal. This could impact the quality of our model’s generated results. For example, in  \cref{pic:failure_case}(a), the wings exhibit stroke disconnection and blurring issues. Efficient hyperparameter search could potentially lead to improvements.
Second, due to the decoupling of shape and position, we occasionally observe that the generated bounding box may fail to cover all strokes, resulting in a ``hard cropping" of the outermost stroke. The strokes in the wings and head in \cref{pic:failure_case}(b) have experienced varying degrees of ``hard cropping". Introducing a coverage regularizer could help address this issue. We leave the two as future work.

\begin{figure}[!h]
    \centering
    \includegraphics[width=0.97\linewidth]{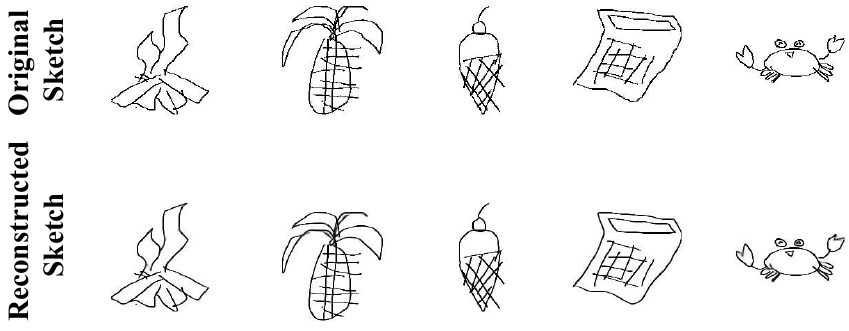}
    \vspace{-2mm}
    \caption{Sketch reconstruction on QuickDraw sketches.}
    \label{pic:quickDraw_recons}
\end{figure}

\parag{Shape codebook generalization.}
\revi{Our learned shape codebook demonstrates strong generalization ability to unseen sketches from other datasets, even without the need for finetuning. This indicates that the learned representations capture fundamental structural patterns of sketches, enabling effective reconstruction across different data distributions.  

Fig. \ref{pic:quickDraw_recons} presents several examples of sketch reconstruction from the QuickDraw dataset \cite{radford2021learning}. As shown in the figure, our shape codebook successfully reconstructs diverse sketches with high fidelity, preserving key structural details and stroke arrangements. The ability to generalize across datasets highlights the robustness of our learned representations, suggesting their potential applicability in various sketch-based tasks, such as sketch recognition, generation, and editing. Furthermore, the high-quality reconstruction results indicate that our codebook effectively captures essential shape priors, making it a versatile tool for sketch-related applications without requiring additional domain-specific adaptations.

}

\parag{Point sequence encoding}
\reviICCV{Following ContextSeg~\cite{wang2024contextseg}, we use stroke bitmap sequences rather than point sequences to represent sketches. The latter yields inferior reconstructions (see Fig.5 in \cite{wang2024contextseg}).
We conducted further statistical analysis in our generation scenario (it is segmentation in ContextSeg). 
Specifically, we replace the CNN with Sketchformer~\cite{ribeiro2020sketchformer} (a Transformer), and train the VQ-VAE and our generator. 
In terms of reconstruction, Sketchformer only yields Acc=0.44/Rec=0.32, which is far below ours (Acc=0.96/Rec=0.97). 
As for the generation, our advantages are more obvious (theirs vs. \textbf{ours}): 75.42 vs. \textbf{15.78} (FID↓), 12.31 vs. \textbf{18.92} (GD↑), 0.16 vs. \textbf{0.53} (CS, close to and $>$0.45 is better), confirming the advantage of bitmap encoding.}

\section{Detailed Network Configuration}

Intuitively, the stroke shape embedding and the learning of the codebook could be achieved simultaneously. However, the network responsible for learning stroke shape embeddings needs to be applied individually to each stroke to effectively capture its structural details, whereas the learning of the codebook must consider the entire sequence of strokes, performing the compression in the sketch level.
Consequently, achieving optimal performance is challenging when attempting to combine these tasks. Therefore, we propose to handle them separately. To this end, we first convert each stroke into a latent embedding and further compress them in the latent space. 

\begin{figure*}[!h]
    \centering
    \includegraphics[width=1\textwidth]{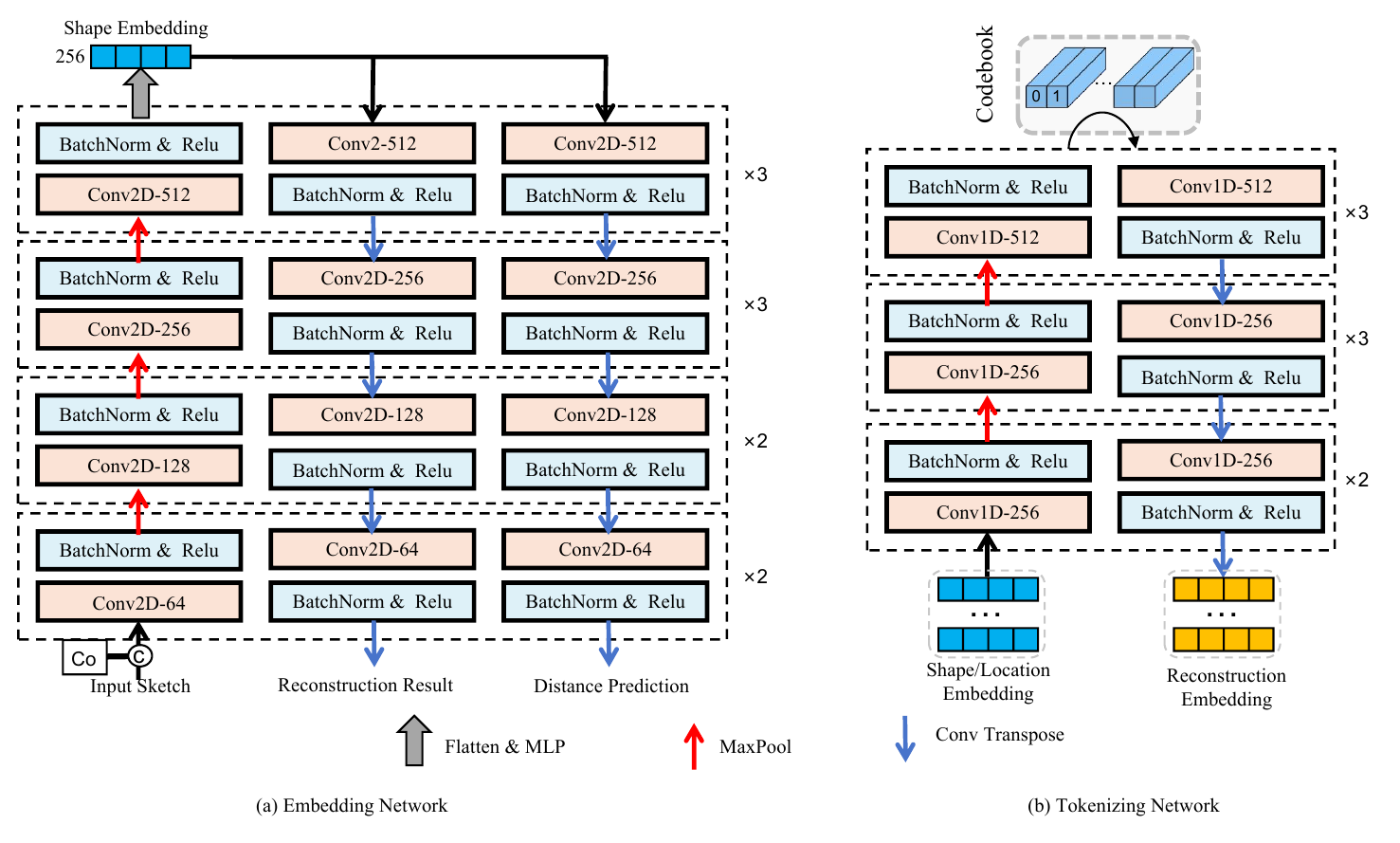}
    \vspace{-7mm}
    \caption{Detailed structure of our embedding networks and tokenizing network.}
    \vspace{-3mm}
    \label{pic:suppl_pipeline}
\end{figure*}

Figure \ref{pic:suppl_pipeline} illustrates the detailed structure of our network. The embedding network has an encoder-decoder structure, accepting the grayscale stroke image input augmented with x and y coordinate channels. Specifically, the encoder comprises a total of 10 layers, which are grouped into four blocks. Each block is characterized by distinct feature dimensions (\ie, 64, 128, 256, and 512, respectively), resulting in a stroke embedding in $\mathbb{R}^{256}$. Both decoder branches share the encoder structure, working symmetrically to transform the stroke embedding into stroke reconstruction and the distance map as in \cite{wang2024contextseg}. 

We also illustrate the detailed network architecture of our tokenizing network, which is used for learning the shape or location codebook. The input is a feature matrix in $\mathbb{R}^{N\times 256}$ composed of shape or location embeddings within a sketch. A Conv1d layer is first employed for feature extraction, followed by a MaxPool layer or ConvTranspose1d layer to compress or expand the feature dimensions. Specifically, the encoder comprises 8 layers grouped into three blocks, each characterized by distinct feature dimensions (\ie, 256, 256, and 512), resulting in a feature of $\mathbb{R}^{N\times 512}$. The ``x3" and ``x2" of each block in \cref{pic:suppl_pipeline} mean that the block is repeated 3 or 2 times. Subsequently, the feature corresponding to each stroke is replaced with the nearest code and then fed into the decoder. Both decoder branches share the encoder structure. The final output aims to reconstruct the latent input.

\end{document}